\documentclass{article}
\usepackage{spconf,amsmath,graphicx}
\usepackage{spconf,amsmath,graphicx}
\usepackage{graphicx}
\usepackage{color}
\usepackage{stfloats}
\usepackage{booktabs}
\usepackage{cite}
\usepackage{amsmath,amssymb,amsfonts}
\usepackage{algorithmic}
\usepackage{graphicx}
\usepackage{textcomp}
\usepackage{xcolor}
\usepackage{url}

\usepackage{fancyhdr}

\title{SEA-Net: Squeeze-and-Excitation Attention Net for Diabetic Retinopathy Grading}
\name{Ziyuan Zhao\thanks{* Both authors contribute equally to this work.}$^{*\dag}$, Kartik Chopra$^{*\ddag}$, Zeng Zeng$^{\sharp \dag}$\thanks{$\sharp$ Corresponding author. The work was supported by Singapore-China NRF-NSFC Grant (Grant No. NRF2016NRF-NSFC001-111).}, Xiaoli Li$^{\dag}$}

\address{$^{\dag}$Institute for Infocomm Research, Agency for Science, Technology and Research, Singapore\\
$^{\ddag}$Institute of Systems Science, National University of Singapore, Singapore
}
\begin{document}
%
\maketitle
\thispagestyle{fancy}
\fancyhead{}
\lhead{}
\vspace{-0.5pt}
\lfoot{\footnotesize{Copyright 2020 IEEE. Published in the IEEE 2020 International Conference on Image Processing (ICIP 2020), scheduled for 25-28 October 2020 in Abu Dhabi, United Arab Emirates. Personal use of this material is permitted. However, permission to reprint/republish this material for advertising or promotional purposes or for creating new collective works for resale or redistribution to servers or lists, or to reuse any copyrighted component of this work in other works, must be obtained from the IEEE. Contact: Manager, Copyrights and Permissions / IEEE Service Center / 445 Hoes Lane / P.O. Box 1331 / Piscataway, NJ 08855-1331, USA. Telephone: + Intl. 908-562-3966.}}
\cfoot{}
\rfoot{}

\begin{abstract}

Diabetes is one of the most common disease in individuals. \textit{Diabetic retinopathy} (DR) is a complication of diabetes, which could lead to blindness. Automatic DR grading based on retinal images provides a great diagnostic and prognostic value for treatment planning. However, the subtle differences among severity levels make it difficult to capture important features using conventional methods. To alleviate the problems, a new deep learning architecture for robust DR grading is proposed, 
referred to as SEA-Net, in which, spatial attention and channel attention are alternatively carried out and boosted with each other, improving the classification performance. In addition, a hybrid loss function is proposed to further maximize the inter-class distance and reduce the intra-class variability. Experimental results have shown the effectiveness of the proposed architecture.

\end{abstract}
\begin{keywords}
Convolutional neural network; Squeeze-and-Excitation net; Diabetic retinopathy grading; Attention mechanism
\end{keywords}

\section{Introduction}
\label{sec:intro}

\textit{Diabetic retinopathy} is the most prevalent microvascular complication among patients with diabetes mellitus~\cite{nentwich2015diabetic}, as well as one of the most frequent cause of blindness of humans~\cite{priya2013diagnosis}. High blood glucose levels can damage the tiny blood vessels at the back of the eyes, even in the prediabetes stage. In Singapore, around 1 out of 12 people aged from $19$ to $69$ years are affected by diabetes, and $43.5\%$ among them suffer from different severity of DR~\cite{website1}. Moreover, there are no early warning symptoms for DR, which lead to difficulties in timely diagnosis and early treatment. Conventionally, DR grading relies on a manual process performed by experts based on fundus photography, which is tedious, costly and time-consuming. Because of human intervention, DR grading also suffers from high intra- and inter-observer variability.



Automatic methods based on computer vision have shown the promising performance in DR detection. Early work in the literature rely on handcrafted features, in which, retinal features such as vessel enhancement, optic disk detection and lesion segmentation are extracted by using image processing techniques and followed by a binary classifier~\cite{sharif2019automatic, pinz1998mapping}. These methods can not be well embedded into an end-to-end framework, and may suffer from the sensitiveness of conditions, like noises and artifacts. Recently, deep learning, especially \textit{convolutional neural networks} have achieved great success in this scenario~\cite{lim2014transformed, wang2015hierarchical, gulshan2016development, SeSeNet2019}. Most of them have focused on binary severity-level classification, while multiple severity-level classification can better help patients and doctors in clinical  diagnosis and treatments. Bravo~\textit{et al.}~\cite{bravo2017automatic} adopted pre-trained CNNs with different image processing techniques to get an average class accuracy of $50.5\%$ on DR grading. BiRA-Net proposed in~\cite{zhao2019bira} utilized the bilinear learning strategy with attention models for fine-grained classification~\cite{lin2015bilinear}, which achieves a higher accuracy. However, even using low-rank bilinear pooling~\cite{kong2017low}, high-dimensional bilinear features also easily lead to some issues, such as overfitting and heavy computation.

In this work, we propose a deep CNN architecture comprising of feature attention and channel recalibration. More specifically, a residual neural network (ResNet)~\cite{resnet} is first implemented to extract features from retinal images, followed by the proposed attention model, which combines a series of $1\times1$ convolution layers for feature refinement in the spatial dimension. To explore the interdependencies between the channels of convolutional features, the \textit{Squeeze-and-Excitation} (SE) block~\cite{hu2018squeeze} is implemented in the proposed architecture for higher quality of representations. Moreover, a combination of the weighted cross entropy loss and center loss is proposed to prevent overtraining, enhancing the discriminative ability of the architecture. Experimental results have demonstrated the effectiveness of the proposed method compared to other methods. 

The remainder of this paper is structured in the following way. We present an overview of the related work in DR grading in Section~\ref{sec:relat}. The proposed method is shown in Section~\ref{sec:methods}. Section~\ref{sec:experiments} shows experimental results and evaluates the proposed method with respect to different metrics. Finally, we present our conclusions in Section~\ref{sec:conclusion}.

\begin{figure*}[t]
    \centering
    \includegraphics[width = 15cm]{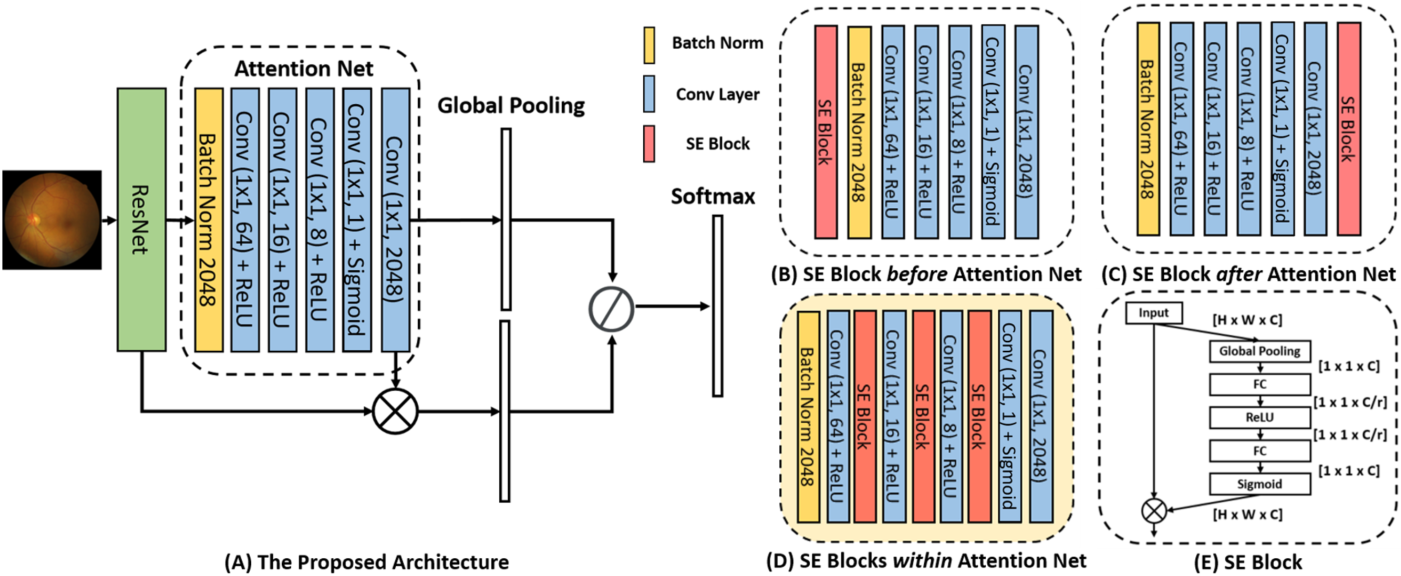}
    \caption{The overview of the proposed framework is shown in (A). The different defined positions of SE block placed in the architecture are illustrated in (B), (C) and (D). The optimal position (D) is highlighted, which is termed as ``SEA-Net". The details of SE block are shown in (E).}
    \label{fig:archi}
\end{figure*}
\section{Related Work}
\label{sec:relat}

\begin{figure}[tb]
    \centering
    \includegraphics[width=0.45\textwidth]{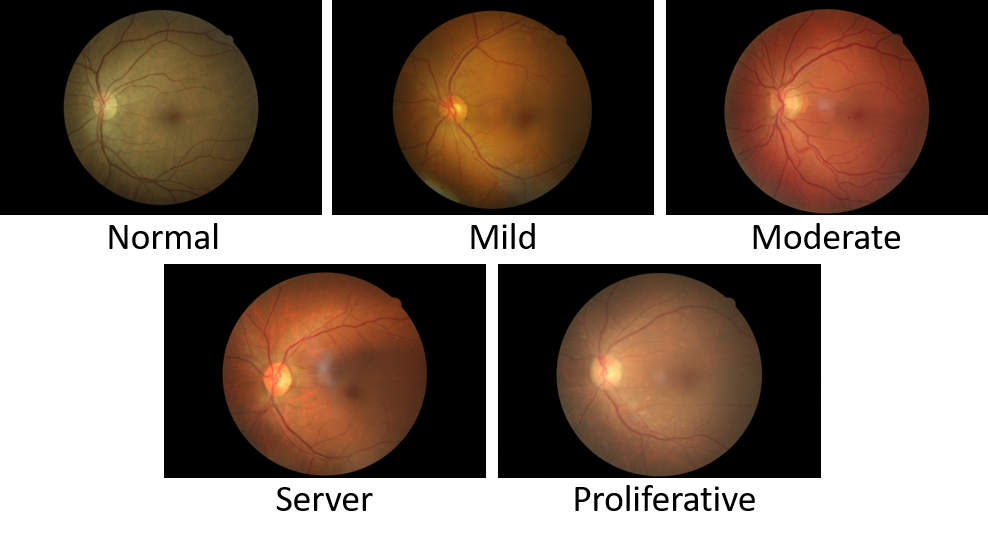}
    \caption{Examples of retinal images provided by EyePACS. The appearance of images from different classes is similar. }
    \label{fig:sample}
\end{figure}

Most automatic DR classification frameworks heavily rely on handcrafted features or extracted features using image processing. In recent years, many work on binary classification of DR have demonstrated the effectiveness of deep learning based methods, while few work focused on multiclass classification of DR, which is more effective for diagnosis and treatments. Bravo~\emph{et al.} integrated different image processing techniques along with VGG16-based architecture for DR grading on a balanced dataset~\cite{bravo2017automatic}. Zeng~\emph{et al.} proposed a siamese neural network for DR grading using left and right retina images~\cite{zeng2019automated}. However, only subtle differences among classes are observed in retina images as Fig.~\ref{fig:sample}, which poses a challenge for DR grading. 

Retina images contain some irrelevant information, and only some features like microaneurysms are critical for doctors. Therefore, to mimic the clinician diagnosis process, Wang~\emph{et al.} proposed a deep learning network using attention mechanism, termed as Zoom-in Net~\cite{wang2017zoom}. In Zoom-in Net, the suspicious areas generated by attention maps are zoomed in for details, and more local information is considered for classifying diabetic retinopathy. In~\cite{zhao2019bira}, a bilinear learning strategy with attention network is proposed to classify DR images at a fine-grained level, boosting meaningful features while suppressing weak ones. These methods capture spatial correlations, achieving competitive accuracy by incorporating spatial attention.

It is well noted that \textit{Squeeze-and-Excitation Networks} (SE Nets) give us a different aspect of attention mechanism, in which, the channel relationship is modeled by introducing SE blocks. The SE Nets perform feature recalibration, utilizing global information for channel attention. 
\section{Methodology}
\label{sec:methods}
The proposed framework is shown in Fig.~\ref{fig:archi}, in which,  a ResNet is firstly trained on the processed images to extract features, followed by Attention Net with a sequence of $1 \times 1$ convolution layers and pooling layers for dimensionality reduction and spatial attention. Considering the channel relationship of learned features, SE blocks are introduced to recalibrate channel-wise feature maps for fine-grained classification. 

\subsection{Residual Neural Network}
\textit{Residual Neural Network} (ResNet)~\cite{resnet} is implemented first for deep feature extraction, in which, the shortcut connections skip some layers and perform identity mapping, thus avoiding the gradient vanishing problem and increasing the training speed and effects. Therefore, the ResNet-50 pre-trained on ImageNet~\cite{deng2009imagenet} is applied as the backbone of the proposed architecture, in which, the weights of all layers are freezed.
\subsection{Attention Net}
Let us assume an input image $\mathbf{I} \in \mathbb{R}^{H \times W \times C^{\prime}}$ which first passes through the ResNet-50 to generate output feature map $\mathbf{U} \in \mathbb{R}^{H \times W \times C}$, $\mathbf{F}_{res}: \mathbf{I} \rightarrow \mathbf{U}$. Here $H$ and $W$ are the spatial height and width, with $C^{\prime}$ and $C$ being the input and output channels, respectively. We define Attention Net as $\mathbf{F}_{atten} = \left[\mathbf{c}_{1}, \mathbf{c}_{2}, \cdots, \mathbf{c}_{n}\right]$, where $\mathbf{c}_{n}$ is n-th convolution layer in Attention Net. Through a series of convolutional layers and non-linearities defined by $\mathbf{F}_{atten}(\cdot)$, the refined feature map $\mathbf{A} \in \mathbb{R}^{H \times W \times C}$ is generated. We consider the feature map $\mathbf{A}=\left[\mathbf{a}_{1}, \mathbf{a}_{2}, \cdots, \mathbf{a}_{C}\right]$ as a combination of channels $\mathbf{a}_{i} \in \mathbb{R}^{H \times W}$. Spatial squeeze is performed by a global average pooling (GAP) layer, which produces vector $\mathbf{x} \in \mathbb{R}^{1 \times 1 \times C}$ with its k-th element

\begin{equation}
x_{k}=\frac{1}{H \times W} \sum_{i}^{H} \sum_{j}^{W} \mathbf{a}_{k}(i, j).
\end{equation}

The GAP layer provides a receptive field of whole spatial extent and embeds the global spatial information in vector $x$. The same operations are performed on $\mathbf{U} \in \mathbb{R}^{H \times W \times C}$, resulting in vector $\mathbf{y} \in \mathbb{R}^{1 \times 1 \times C}$. Finally, to filter unrelated information, an element-wise division is used followed by a softmax layer.

\subsection{Squeeze-and-Excitation Block}
To exploit channel dependencies and contextual information, we propose to incorporate the SE block~\cite{hu2018squeeze} into the proposed architecture. The details of Squeeze-and-Excitation block is described in Fig.~\ref{fig:archi} (E). The input $\mathbf{G} \in \mathbb{R}^{H \times W \times C}$ is first embedded in to vector $\mathbf{z} \in \mathbb{R}^{1 \times 1 \times C}$ using a GAP layer, and then transformed to $\hat{\mathbf{z}} = \mathbf{W}_{1}\left(\delta\left(\mathbf{W}_{2} \mathbf{z}\right)\right)$, where $\mathbf{W}_{1} \in \mathbb{R}^{C \times \frac{C}{r}}$ and  $\mathbf{W}_{2} \in \mathbb{R}^{\frac{C}{r} \times C}$ are the weights of two fully-connected (FC) layers, and $\delta(\cdot)$ is the ReLU activation layer. The parameter $r$ is the reduction ratio for dimensionality reduction, indicating the bottleneck in the channel excitation. In our experiments, we set $r = 4$. After passing $\hat{\mathbf{z}}$ through a sigmoid layer $\sigma(\hat{\mathbf{z}})$, the activations of $\hat{\mathbf{z}}$ are limited into the interval $[0,1]$, which is used to recalibrate the input $\mathbf{G}=\left[\mathbf{g}_{1}, \mathbf{g}_{2}, \cdots, \mathbf{g}_{C}\right]$, $\mathbf{g}_{i} \in \mathbb{R}^{H \times W}$. The output feature map $\mathbf{G}_{se} \in \mathbb{R}^{H \times W \times C}$, $\mathbf{F}_{se}: \mathbf{G} \rightarrow \mathbf{G}_{se}$ is computed as
\begin{equation}
\mathbf{G}_{se}=  \left[\sigma\left(\hat{z}_{1}\right) \mathbf{g}_{1}, \sigma(\hat{z_{2}}) \mathbf{g}_{2}, \cdots, \sigma(\hat{z_{C}}) \mathbf{g}_{C}\right].
\end{equation}

When integrating the SE blocks in the proposed architecture, the positions of SE blocks in the network influence the performance of DR grading. To find the optimal position, we explore three different positions of SE blocks. Fig.~\ref{fig:archi} (B), (C) and (D) show different defined positions of SE blocks.

\subsection{The Proposed Hybrid Loss}
Considering the distance between different classes, we propose to implement center loss~\cite{wen2016discriminative} to reduce the loss-accuracy discrepancy and get an improved convergence. The center loss function is formulated as 

\begin{equation}
\mathfrak{L}_{ct}=\frac{1}{2} \sum_{i=1}^{m}\left\|\boldsymbol{x}_{i}-\boldsymbol{c}_{y_{i}}\right\|_{2}^{2},
\end{equation}

where $x_{i}$ is the $i^{th}$ training sample and $c_{yi}$ is the $y^{th}$ class center of deep features. 

In addition, the DR dataset is heavily imbalanced, and most images are labeled as 0. The weighted cross entropy loss is used to alleviate the problem, which is defined by


\begin{equation}
\mathfrak{L}_{ce}=\operatorname{weight}_{y}\left(-\log \left(\frac{\exp (x[y])}{\sum_{j} \exp (x[j])}\right)\right),
\end{equation}


where $x$ is the training sample, $y\in \left [ 0,C-1 \right ]$, $C$ is the number of classes, $\operatorname{weight}_{y}$ is a manual rescaling weight given to each class. The $\operatorname{weight}_{y}$ is calculated by dividing the total number of training samples by samples in each class. This ensures that minority classes get higher weights.

Finally, the hybrid loss function is denoted as
\begin{equation}
    \mathfrak{L} = \mathfrak{L}_{ce} + \lambda\mathfrak{L}_{ct},
\end{equation}

where $\lambda$ is a scalar to control the strength of loss functions. It is noted that when $\lambda = 0$, only the weighted cross entropy loss is used for parameter estimation.






\section{Experiments}
\label{sec:experiments}

\subsection{Dataset and implementation}
The dataset used in this work is from a Kaggle competition provided by EyePACS~\cite{kaggle}. The dataset is comprised of $33566$ retinal images with 5 classes, where 0 represents no disease and 4 represents the highest severity of disease. Some examples are shown in Fig.~\ref{fig:sample}. Due to the imbalanced dataset, following the data distribution adopted in~\cite{bravo2017automatic}, a balanced testing dataset of 1560 images was applied to our experiments for testing, and the rest were used for training.

To reduce the noises from background, the original images are cropped and resized to $610\times610$ so that only the retinal region is visible. Then the images are standardized across RGB channels by subtracting the mean and divided by the standard deviation of each channel in the training data. The histogram equalization is applied to improve contrast of the images. To reduce overfitting on models, some augmentation techniques are carried on the dataset, more specifically, the images are randomly rotated  $\pm$10 degrees, flipped vertically or horizontally.

The proposed framework is trained on a single NVIDIA Quadro P5000 GPU with batch size of 20, using the stochastic gradient descent (SGD) optimizer with a momentum of 0.9. The initial learning rate is $0.002$ with weight decay factor of $1\times e^{-8}$.

\subsection{Performance metrics}
To evaluate the performance on the multi-class classification task, we generate the confusion matrix, in which, the number of predictions in each class are presented. Based on the confusion matrix, the \textit{average of classification accuracy} (ACA) can be calculated by taking the percentage of the diagonal elements, which represent the number of points for which the predicted label is equal to the true label. By adopting one-against-all strategy~\cite{bishop2006pattern}, F1 score of each class can be calculated, and Marco-F1 score is obtained by taking the average of the F1 scores for all 5 classes.

To evaluate the diagnostic performance of the proposed method, Receiver operating characteristics (ROC) ~\cite{hajian2013receiver} are calculated, which describe the true positive rate (TPR) and the false positive rate (FPR) at various threshold settings. For comparison, AUC (Area Under Curve) is calculated based on ROC.


\begin{table}[t]
\caption{Experimental results of various approaches on DR grading, which are described in Section~\ref{subsec:baseline}. The last one is implemented with the proposed hybrid loss function, and the results are obtained when $\lambda = 0.1$.}
\label{table_results}
\begin{center}
\begin{tabular}{cccc}
\hline
Method               & ACA    & Marco-F1 & AUC \\ \hline
Bravo et al.~\cite{bravo2017automatic} & 0.5051 & 0.5081   & -   \\
BiRA-Net ~\cite{zhao2019bira}    & 0.5431 & 0.5725   & -   \\
AT-Net               & 0.5442      & 0.4951        &  0.8699  \\
SE-AT-Net            & 0.5776      & 0.5505        &  0.8734   \\
AT-SE-Net            & 0.5830      & 0.5892        & 0.8721    \\
SEA-Net            & 0.5859        & 0.5872        & 0.8738   \\ 
SEA-Net ($\lambda = 0.1$)            &  \textbf{0.5994}      &  \textbf{0.6047}        &  \textbf{0.8760}    \\ \hline
\end{tabular}
\end{center}
\end{table}

\subsection{Baseline methods}
\label{subsec:baseline}
Previous deep learning approaches are described for comparison. In~\cite{bravo2017automatic}, a series of VGG-based classifiers were trained using different image processing techniques, such as circular RGB and color centered sets. In~\cite{zhao2019bira}, a bilinear learning strategy was proposed to improve the classification performance on fine-grained images. In addition, the proposed architecture with different positions of SE block is implemented as depicted in Fig.~\ref{fig:archi}, which are summarized as follows.

\begin{itemize}
\vspace{-9pt}
	\item \textbf{AT-Net:} The proposed Attention Net.
	\vspace{-7pt}
    \item \textbf{SE-AT-Net:} Net with SE block \textit{before} attention block.
    \vspace{-19pt}
    \item \textbf{AT-SE-Net:} Net with SE block \textit{after} attention block.
    \vspace{-7pt}
    \item \textbf{SEA-Net:} Net with SE blocks \textit{within} attention block.
\end{itemize}

\subsection{Results and discussion}

The experimental results of all methods on the testing data are shown in Table~\ref{table_results}. We can see that the proposed framework outperforms other methods in all metrics, even without SE block. SE blocks are proved to be effective in the proposed methods, in which, the best results are obtained from SEA-Net, which is the optimal position in the network. In SEA-Net, the SE blocks are placed alternatively with convolution layers, recalibrating the learned feature maps in an adaptive manner. 
 
\begin{figure}[tb]
    \centering
    \includegraphics[width=8.5cm]{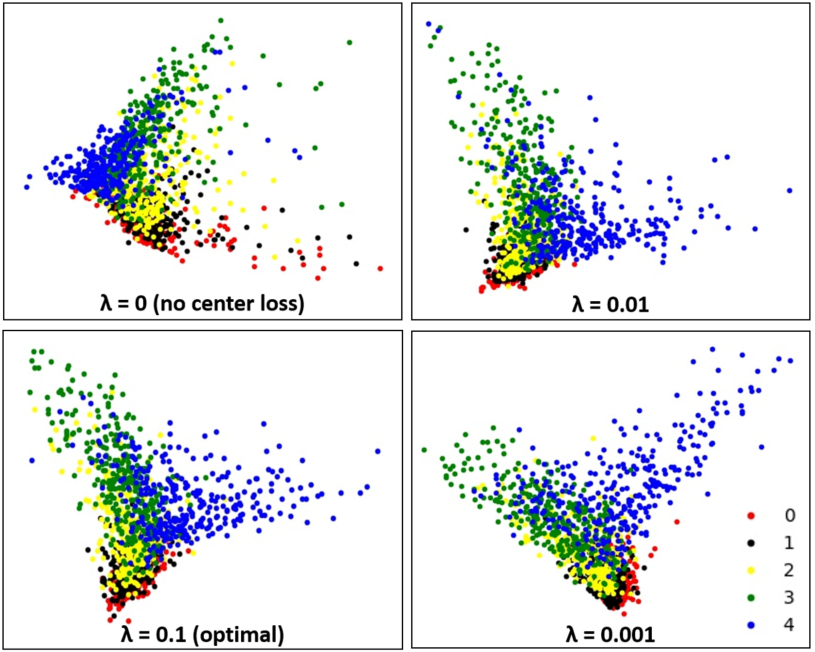}
    \caption{The feature distribution learned with the proposed hybrid loss of different $\lambda$ values. The points with different colors denote features from different classes.}
    \label{fig:visual}
\end{figure}

It is well noted that the performance of SEA-Net is further improved with the proposed hybrid loss function. Following~\cite{wen2016discriminative}, the distribution of deeply learned features is visualized in Fig.~\ref{fig:visual}, which proves that the proposed hybrid loss can learn better discriminative features, especially for confusing classes, {\textit{i}.\textit{e}.}, class 0 and class 1. These two classes are usually regarded as one class in binary classification~\cite{bravo2017automatic}. In addition, the features in the same class are more close under the supervision of the hybrid loss, than only using the weighted cross entropy loss.


\vspace{-14pt}
\section{Conclusions}
\label{sec:conclusion}

In this work, we proposed a novel deep learning architecture for DR grading, called ``SEA-Net", in which, spatial attention and channel attention are implemented to boost each other, recalibrating the attention maps adaptively. In addition, to obtain better learned features, a hybrid loss function based on weighted cross entropy loss and center loss is implemented in SEA-Net. Extensive experiments were carried out using different methods, which demonstrate the effectiveness of the proposed architecture.



\clearpage
\bibliographystyle{IEEEbib}
\bibliography{refs}

\end{document}